# Deep Neural Networks for Swept Volume Prediction Between Configurations

Hao-Tien Lewis Chiang[1]    Aleksandra Faust[2]    Lydia Tapia[1]

*Abstract*— Swept Volume (SV), the volume displaced by an object when it is moving along a trajectory, is considered a useful metric for motion planning. First, SV has been used to identify collisions along a trajectory, because it directly measures the amount of space required for an object to move. Second, in sampling-based motion planning, SV is an ideal distance metric, because it correlates to the likelihood of success of the expensive local planning step between two sampled configurations. However, in both of these applications, traditional SV algorithms are too computationally expensive for efficient motion planning. In this work, we train Deep Neural Networks (DNNs) to learn the size of SV for specific robot geometries. Results for two robots, a 6 degree of freedom (DOF) rigid body and a 7 DOF fixed-based manipulator, indicate that the network estimations are very close to the true size of SV and is more than 1500 times faster than a state of the art SV estimation algorithm.

## I. INTRODUCTION

Swept Volume (SV) is the volume displaced by an object when it is moving along a trajectory [3], [1]. Essentially, it is the union of the volumes of all configurations of the object along a trajectory. Computing SV requires computing a complex geometry in an often high-dimensional configuration space (C-space), where each point in this space completely describes the robot geometry. SV is useful in many applications such as geometric modeling [6], robot workspace analysis [2], collision avoidance [11] and motion planning [16].

SV has been identified as being particularly useful for robot motion planning since the performance of sampling-based motion planners, such as probabilistic roadmap (PRM) [10] and rapidly exploring random tree (RRT) [14], depends greatly on a distance metric that returns an estimated distance between two sampled configurations [4]. The distance metric determines the configuration pairs that are selected for the expensive local planning operation that makes roadmap connections in a PRM or tree extensions in RRT. Intuitively, a good metric should limit the connect or extend operations to those that are most likely to succeed, i.e., free of collision [4]. The size of SV between configurations has been identified as an ideal distance metric [3] since it is related to the probability of collision between two points in C-space [13].

Computation of exact SVs is intractable, and all practical SV algorithms focus on generating an approximate SV [12]. Common approaches for computing approximate SVs include occupation grid-based and boundary-based methods. Occupation grid-based approaches decompose the workspace, e.g., into voxels, in order to record the robot's workspace occupation as it executes a trajectory [8], [17]. The resulting approximation of SV has a resolution-tunable accuracy and is conservative, which can be critical for applications such as collision avoidance [15]. The boundary-based methods identify and record the polygons contributing to the outer most boundary of the SV. These polygons are then used to extract the boundary surface [5], [12], [3]. Despite advances in approximate SV computation, SV is still considered too expensive to be used as the distance metric in sampling-based motion planners [13].

In this paper, Deep Neural Networks (DNNs) learn the complex and nonlinear relationship between trajectories in C-space and the corresponding size of SV for a variety of robot geometries. The trained DNNs can quickly return the estimated size of SV between any pair of configurations. To train the networks, we generate training data by randomly sampling pairs of configurations and computing the approximate SV between each pair using an occupation grid-based method [17]. The DNNs are then trained to infer the size of SV in a supervised manner.

To evaluate the quality of SV learning, we trained and evaluated two DNNs for two robot types, a six degree of freedom (DOF) rigid body and a 7 DOF fixed-based manipulator. While each DNN was trained independently, the two DNNs were trained using the same hyper-parameters, e.g., structure of hidden layers, training batch size, learning rate and number of epochs. Results indicate that these networks can accurately estimate the size of SV and are more than 1500 times faster than a state of the art approximate SV computation algorithm.

## II. METHOD

The size of SV for a trajectory in C-space can be described by a function $\mathcal{SV}(c_1, c_2)$, where $c_1$ and $c_2$ are the start and end configurations. In this work, we assume the trajectory between $c_1$ and $c_2$ is a straight line in C-Space. $\mathcal{SV}(c_1, c_2)$ can be highly complex and nonlinear due to rotational degrees of freedoms, especially in cases where the robot has an articulated body. To approximate this complex function, we utilize a simple fully-connected feedforward DNN, which has been shown to be able to approximate any continuous bounded function [9]. Training data is generated with an octree occupation grid-based method. We explain the DNN and training data generation below.

[1] Chiang and Tapia are with Department of Computer Science, University of New Mexico, Albuquerque, NM, USA. e-mail: lewispro,tapia@cs.unm.edu

[2] Faust is with Google Brain, Mountain View, CA 94043, USA e-mail: faust@google.com.

[3] Formally, SV would be a distance semimetric. However, we use the planning terminology distance metric in this work.

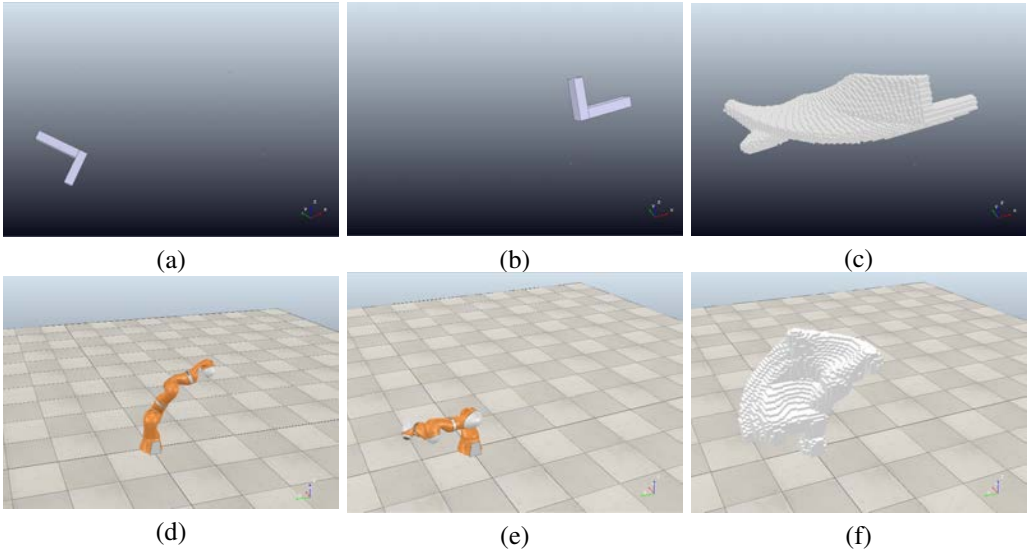

Fig. 1: Example *start* (a, d) and *end* (b, e) configurations and the corresponding $\mathcal{SV}(start, end)$ for a L-shaped rigid-body robot (top row) and a fixed-base Kuka LBR IIWA 14 R820 manipulator (bottom row).

### A. Generating Training Data

The training data is composed of many pairs of $c_1$ and $c_2$ and the corresponding $\mathcal{SV}(c_1, c_2)$. Each pair of $c_1$ and $c_2$ is randomly sampled from C-space. To compute $\mathcal{SV}(c_1, c_2)$, we implemented a state of the art octree-based SV algorithm [17], where the trajectory of the robot is represented by $N$ intermediate C-space configurations. Since we only consider straight line C-space trajectories, the $n^{th}$ intermediate configuration is

$$c_n = (1 - n/N)c_1 + n/N c_2. \quad (1)$$

Next, the forward kinematics of the robot maps each $c_n$ to the workspace occupancy of the robot

$$\mathcal{G}_n(x,y,z) = \begin{cases} 1, & \text{robot overlaps with point } (x,y,z) \\ 0, & \text{otherwise.} \end{cases} \quad (2)$$

The SV can be approximated by taking the union of all $\mathcal{G}_n$. The occupancy and the union operation can be approximated by an octree decomposition of workspace up to a resolution, $\Delta$, in order to speed up computation compared to an uniformly distributed voxel grid. Lastly, $\mathcal{SV}(c_1, c_2)$ can be computed by adding the volume of all occupied cubes in the octree.

### B. Learning SV

We use a simple fully-connected feedforward DNN (Figure 2) to approximate $\mathcal{SV}(c_1, c_2)$. Our network has $2M$ input neurons where $M$ is the dimension of the C-space. The first $M$ input neurons have activations equal to the components of $c_1$ while the second $M$ equals to the components of $c_2$. The input layer is connected to the hidden layers of $k$ layers each with $N_i$ neurons using the ReLu [7] activation function. The output layer has one neuron and the amount of activation is the estimated size of SV $\mathcal{SV}'(c_1, c_2)$.

The goal of the network is to learn $\mathcal{SV}(c_1, c_2)$. Therefore, we define the loss function as

$$\text{Loss} = (\mathcal{SV}'(c_1, c_2) - \mathcal{SV}(c_1, c_2))^2. \quad (3)$$

Stochastic gradient descent-based back-propagation algorithms can be used to adjust network parameters to minimize the loss function over a batch of training data.

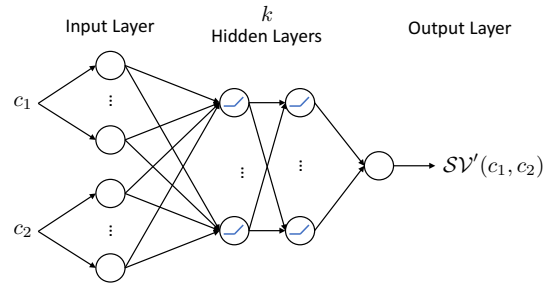

Fig. 2: The feedforward DNN used to compute $\mathcal{SV}'(c_1, c_2)$. $c_1$ and $c_2$ are the start and end configurations of the trajectory, respectively. Neurons in the hidden layers use the ReLu activation function represented by the blue curves.

## III. RESULTS

To evaluate the quality of SV learning, we trained a DNN to learn $\mathcal{SV}$ for each robot. The performance of the network was evaluated by an evaluation data set. This data set was generated in the same fashion as the training data, but it was previously unseen by the network.

The two networks share the same hyper-parameters. These include: the number of hidden layers $k = 3$, the number of neurons in the hidden layers = [1024, 512, 256], learning rate = 0.1, training batch size = 100 and the number of training epochs = 500 (the number of times the network utilizes the entire training data set during training). One hundred thousand training samples and ten thousand evaluation samples were generated for each robot. $N = 100$ intermediate configurations are generated between $c_1$ and $c_2$. The octree resolution was $\Delta = 0.025$m.

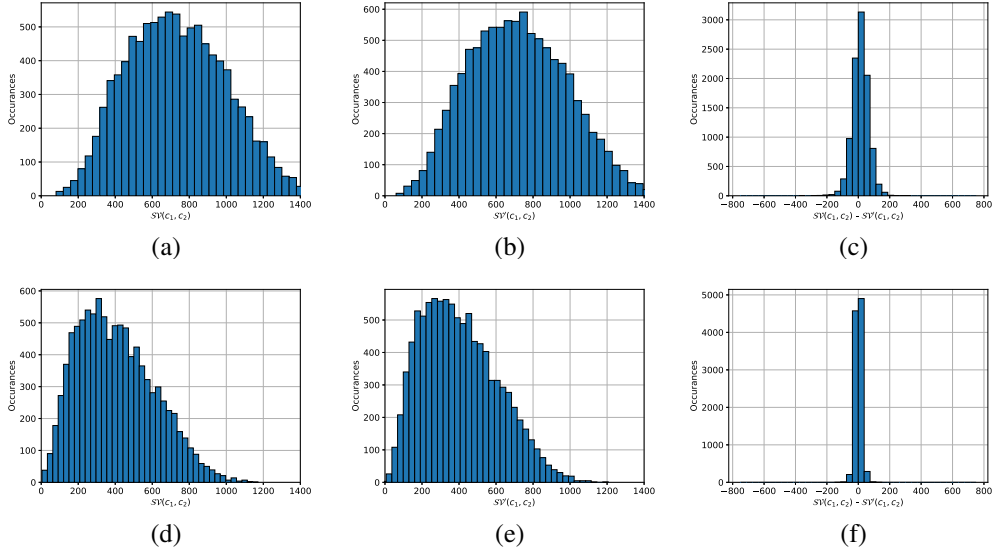

Fig. 3: The size of SV ($\mathcal{SV}(c_1, c_2)$) of an L-shaped robot (a) and a Kuka LBR IIWA 14 R820 (d). The size of SV estimated by the DNN ($\mathcal{SV}'(c_1, c_2)$) for the two robots are shown in (b) and (e). The difference between $\mathcal{SV}'(c_1, c_2)$ and $\mathcal{SV}(c_1, c_2)$ are shown in (c) and (f).

The DNNs are implemented with Tensorflow in Python on an Intel i7-6820HQ at 2.7GHz with 16GB of RAM. The training data generation is implemented within the open-source V-REP robot simulator platform.

The robots are shown in Figure 1. The first is a 6 DOF L-shaped rigid body of size 40cm, 60cm, 10cm (width, height, depth) (shown in Figure 1 (a)). A configuration of the robot is described by the center of mass position and the yaw, pitch and roll of the robot. The second robot is a fixed-based Kuka LBR IIWA 14 R820 (shown in Figure 1 (d)). The seven joint angles describe a configuration of the robot. Configurations are uniform-randomly sampled from [-1.5m, 1.5m] for position axes and [-$\pi$, $\pi$] for rotation axes.

Figures 3 (a) and 3 (d) show the distribution of $\mathcal{SV}(c_1, c_2)$ of ten thousand pairs of $c_1$ and $c_2$ for the evaluation data. Note the differences in distribution between the robot geometries. Figures 3 (b) and 3 (e) show $\mathcal{SV}'(c_1, c_2)$ for the same data as estimated by DNNs. Across robot geometries, there are striking similarities between $\mathcal{SV}'(c_1, c_2)$ and $\mathcal{SV}(c_1, c_2)$ indicating successful learning. In addition, Figure 3 (c) and (f) shows the estimation error ($\mathcal{SV}'(c_1, c_2) - \mathcal{SV}(c_1, c_2)$) which is small, symmetric and centered around zero.

We further explored the predictions of the DNNs by comparing $\mathcal{SV}(c_1, c_2)$ against $\mathcal{SV}'(c_1, c_2)$ and the Euclidean distance (Figure 4). The blue diamonds in Figure 4 show $\mathcal{SV}'(c_1, c_2)$. They closely track $\mathcal{SV}(c_1, c_2)$. For comparison, the green circles show the Euclidean distance between $c_1$ and $c_2$, a commonly used distance metric for sampling-based planners [4]. In order to represent $\mathcal{SV}(c_1, c_2)$, we scale the value such that the average Euclidean distance matches the average $\mathcal{SV}(c_1, c_2)$ of the evaluation data. It it clear that the Euclidean distance does not correlate to $\mathcal{SV}(c_1, c_2)$ well, especially when $\mathcal{SV}(c_1, c_2)$ is large. Since larger $\mathcal{SV}(c_1, c_2)$ implies a higher probability of collision between $c_1$ and $c_2$, an overestimation of $\mathcal{SV}(c_1, c_2)$ misses opportunities to make connections that are likely to succeed. In the opposite case of underestimation, the planner can waste computation attempting connections that are unlikely to succeed.

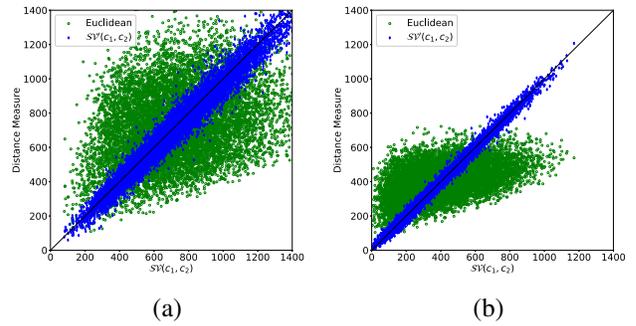

Fig. 4: The size of SV ($\mathcal{SV}(c_1, c_2)$) and the distance measure estimated by the DNN (blue diamonds) and Euclidean C-space distance (green circles) for a L-shaped robot (a) and a Kuka LBR IIWA 14 R820 (b).

| Robot | Training Sample | DNN Training | $\mathcal{SV}'(c_1, c_2)$ with DNN |
|---|---|---|---|
| L-Shaped | 0.46s | 3751.03s | 297.33±6.67$\mu$s |
| Kuka LBR | 4.06s | 4023.35s | 276.97±8.15$\mu$s |

TABLE I: Computation time costs for generating a training data sample, training the DNNs, and estimating $\mathcal{SV}'(c_1, c_2)$ via DNNs.

Table I shows the computation time for generating a training data sample, training the DNNs, and using the DNN to output $\mathcal{SV}'(c_1, c_2)$. Recall that training sample generation involves computing the SV between two configurations and then compute the size of SV. This takes about half a second for the simpler rigid-body geometry and about 4 seconds for the more complex manipulator geometry. This further demonstrates the infeasibility of state of the art SV computation methods as a distance metric for motion planning. On the other hand, estimating the size of SV by DNN is extremely fast. While about 44 times slower than computing the euclidean distance, it is more than 1500 times faster than the state of the art approximated SV.

## IV. CONCLUSIONS AND FUTURE WORK

We demonstrated that a simple DNN can be trained to estimate the size of SV well. In addition, estimating the size of SV from the network is very fast. These facts suggest that a trained DNN for SV can be used as a distance measure for sampling-based motion planners.

We plan to extend our experiments to include more robot types such as mobile manipulators. In addition, we will integrate our method with sampling-based motion planners such as RRT and PRM in order to evaluate the performance gain of using $\mathcal{SV}'(c_1, c_2)$ as a distance metric.

## V. ACKNOWLEDGMENTS

Tapia and Chiang partially supported by the National Science Foundation under Grant Numbers IIS-1528047 and IIS-1553266 (Tapia, CAREER). Any opinions, findings, and conclusions or recommendations expressed in this material are those of the authors and do not necessarily reflect the views of the National Science Foundation.

## REFERENCES

[1] K. Abdel-Malek, J. Yang, D. Blackmore, and K. Joy. Swept volumes: foundation, perspectives, and applications. *International Journal of Shape Modeling*, 12(01):87–127, 2006.
[2] S. Abrams and P. K. Allen. Swept volumes and their use in viewpoint computation in robot work-cells. In *Proc. of IEEE International Symposium on Assembly and Task Planning*, pages 188–193, 1995.
[3] S. Abrams and P. K. Allen. Computing swept volumes. *The Journal of Visualization and Computer Animation*, 11(2):69–82, 2000.
[4] N. M. Amato, O. B. Bayazit, L. K. Dale, C. Jones, and D. Vallejo. Choosing good distance metrics and local planners for probabilistic roadmap methods. In *Proc. IEEE Int. Conf. Robot. Autom. (ICRA)*, volume 1, pages 630–637, 1998.
[5] M. Campen and L. Kobbelt. Polygonal boundary evaluation of minkowski sums and swept volumes. In *Computer Graphics Forum*, volume 29, pages 1613–1622, 2010.
[6] J. Conkey and K. I. Joy. Using isosurface methods for visualizing the envelope of a swept trivariate solid. In *Proc. of IEEE Pacific Conference on Computer Graphics and Applications*, pages 272–280, 2000.
[7] R. H. Hahnloser, R. Sarpeshkar, M. A. Mahowald, R. J. Douglas, and H. S. Seung. Digital selection and analogue amplification coexist in a cortex-inspired silicon circuit. *Nature*, 405(6789):947, 2000.
[8] J. C. Himmelstein, E. Ferre, and J.-P. Laumond. Swept volume approximation of polygon soups. *IEEE Trans. on Autom. Sci. and Eng.*, 7(1):177–183, 2010.
[9] K. Hornik. Approximation capabilities of multilayer feedforward networks. *Neural networks*, 4(2):251–257, 1991.
[10] L. Kavraki, P. Svestka, J. claude Latombe, and M. Overmars. Probabilistic roadmaps for path planning in high-dimensional configuration spaces. In *Proc. IEEE Int. Conf. Robot. Autom. (ICRA)*, pages 566–580, 1996.
[11] J. Kieffer and F. Litvin. Swept volume determination and interference detection for moving 3-D solids. *Journal of Mechanical Design*, 113(4):456–463, 1991.
[12] Y. J. Kim, G. Varadhan, M. C. Lin, and D. Manocha. Fast swept volume approximation of complex polyhedral models. *Computer-Aided Design*, 36(11):1013–1027, 2004.
[13] J. J. Kuffner. Effective sampling and distance metrics for 3d rigid body path planning. In *Proc. IEEE Int. Conf. Robot. Autom. (ICRA)*, volume 4, pages 3993–3998, 2004.
[14] S. M. LaValle and J. J. Kuffner. Randomized kinodynamic planning. *Int. J. Robot. Res.*, 20(5):378–400, 2001.
[15] N. Perrin, O. Stasse, L. Baudouin, F. Lamiraux, and E. Yoshida. Fast humanoid robot collision-free footstep planning using swept volume approximations. *IEEE Trans. Robot.*, 28(2):427–439, 2012.
[16] F. Schwarzer, M. Saha, and J.-C. Latombe. Exact collision checking of robot paths. In *Proc. Int. Workshop on Algorithmic Foundations of Robotics (WAFR)*, pages 25–41. 2004.
[17] A. Von Dziegielewski, M. Hemmer, and E. Schömer. High precision conservative surface mesh generation for swept volumes. *IEEE Trans. on Autom. Sci. and Eng.*, 12(1):183–191, 2015.